\documentclass[10pt,twocolumn,letterpaper]{article}

\usepackage[pagenumbers]{iccv}      

\newcommand{\gr}[1]{{\textcolor{gray}{#1}}}

\definecolor{iccvblue}{rgb}{0.21,0.49,0.74}
\usepackage[pagebackref,breaklinks,colorlinks,allcolors=iccvblue]{hyperref}
\usepackage{multirow}

\def\paperID{*****} 
\def\confName{ICCV}
\def\confYear{2025}

\title{WISA: World Simulator Assistant for Physics-Aware Text-to-Video Generation}

\author{Jing Wang$^{1,2*}$, Ao Ma$^{2*}$, Ke Cao$^{2*}$, Jun Zheng$^{1}$, Zhanjie Zhang$^{2}$, Jiasong Feng$^{2}$,\\  Shanyuan Liu$^{2}$, Yuhang Ma$^{2}$, Bo Cheng$^{2}$, Dawei Leng$^{2\dag}$, Yuhui Yin$^{2}$, Xiaodan Liang$^{1,3\dag}$\\
{\tt\small $^*$Equal Contribution, $^\dag$Corresponding Authors}\\
$^1$Shenzhen Campus of Sun Yat-Sen University, $^2$360 AI Research, $^3$Peng Cheng Laboratory, \\
{\tt\small wangj977@mail2.sysu.edu.cn, lengdawei@360.cn, xdliang328@gmail.com}
}

\begin{document}

\twocolumn[{%
\renewcommand\twocolumn[1][]{#1}%
\maketitle
\vspace{-9mm}
\begin{center}
    \centering
    \includegraphics[width=\linewidth]{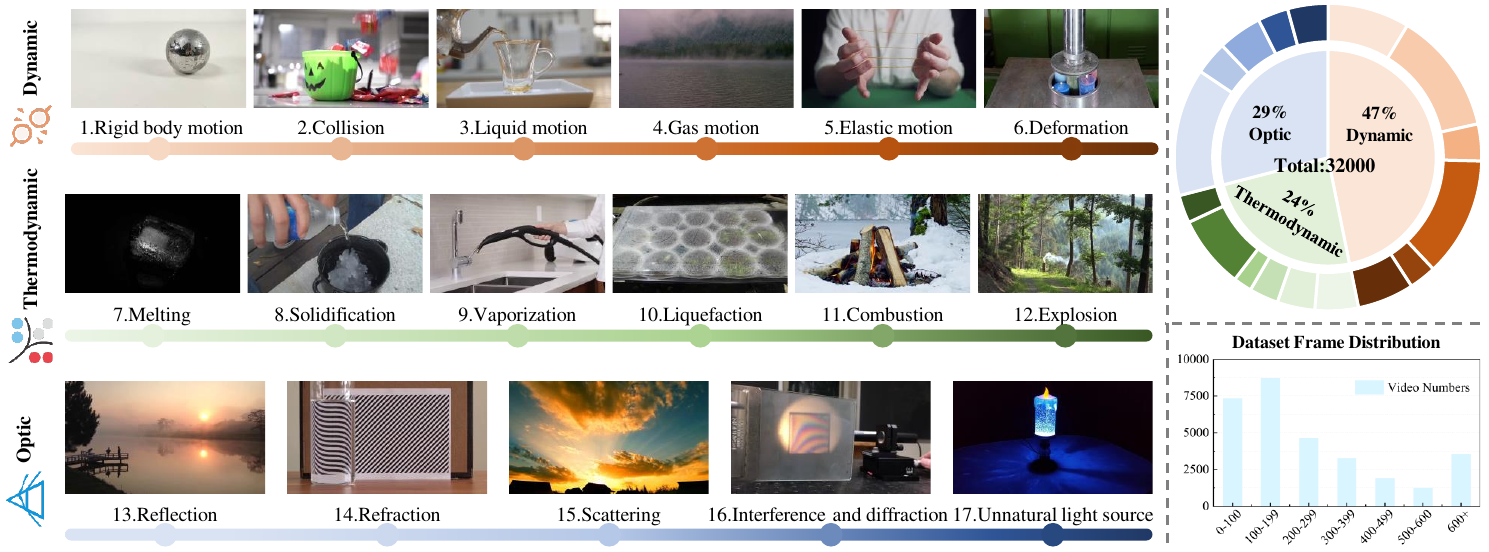}
    \vspace{-6mm}
    \captionof{figure}{
    \textbf{Overview of our physical dataset WISA-32K}. (Left) Examples of 17 physical phenomena across 3 physics categories in WISA-32K. (Top right) WISA-32K contains of approximately 32,000 video clips, with 47\% related to \textit{Dynamics}, 24\% to \textit{Thermodynamics}, and 29\% to \textit{Optics}. (Bottom right) Distribution of frame counts across all videos in WISA-32K.
    } \vspace{1mm}
    \label{fig:intro}
\end{center}%
}]

\begin{abstract}
Recent rapid advancements in text-to-video (T2V) generation, such as SoRA and Kling, have shown great potential for building world simulators. However, current T2V models struggle to grasp abstract physical principles and generate videos that adhere to physical laws. This challenge arises primarily from a lack of clear guidance on physical information due to a significant gap between abstract physical principles and generation models. To this end, we introduce the \textbf{W}orld \textbf{S}imulator \textbf{A}ssistant (\textbf{WISA}), an effective framework for decomposing and incorporating physical principles into T2V models. Specifically, WISA decomposes physical principles into textual physical descriptions, qualitative physical categories, and quantitative physical properties. To effectively embed these physical attributes into the generation process, WISA incorporates several key designs, including Mixture-of-Physical-Experts Attention (MoPA) and a Physical Classifier, enhancing the model's physics awareness. Furthermore, most existing datasets feature videos where physical phenomena are either weakly represented or entangled with multiple co-occurring processes, limiting their suitability as dedicated resources for learning explicit physical principles. We propose a novel video dataset, \textbf{WISA-32K}, collected based on qualitative physical categories. It consists of 32,000 videos, representing 17 physical laws across three domains of physics: dynamics, thermodynamics, and optics. Experimental results demonstrate that WISA can effectively enhance the compatibility of T2V models with real-world physical laws, achieving a considerable improvement on the VideoPhy benchmark. The visual exhibitions of WISA and WISA-32K are available in the \href{https://360cvgroup.github.io/WISA/}{Project Page}.
\end{abstract}    
\vspace{-15pt}
\section{Introduction}
\label{sec:intro}
Many recent studies (e.g., Cosmos \cite{agarwal2025cosmos}, Kling \cite{kling2024videogeneration}, Step-Video-T2V \cite{ma2025step}, Sora \cite{openai2024videogeneration}, and CogVideoX \cite{yang2024cogvideox}) have endeavored to develop robust text-to-video (T2V) models for building world simulators \cite{xiang2024pandora, videoworldsimulators2024, zhu2024sora}. While these models are capable of generating highly realistic and text-consistent videos, leveraging the scale of their data and architectures, they still face challenges in understanding abstract physical principles and producing videos that fully align with real-world physical laws \cite{bansal2024videophy, meng2024towards}.

The substantial gap between abstract physical laws and their visual manifestations presents a significant challenge for injecting physical guidance into T2V models. Physical principles or laws are often conveyed through abstract natural language, reflecting the underlying operational logic of the real world. 
In contrast, generative models map textual descriptions directly to the visual appearance of objects, including their color and shape.  There is a complex logical reasoning process between physical principles and the visual physical phenomena they give rise to.  However, generative models, which are trained to map learned data distributions, struggle to extract appropriate physical information from a single textual instruction and translate it into a physically consistent visual representation for a specific scenario. This challenge becomes even more pronounced in video generation, where the strict temporal order of physical events must be preserved.  

To this end, we propose the \textbf{W}orld \textbf{S}imulator \textbf{A}ssistant (\textbf{WISA}), which decomposes abstract physical principles into multiple categories of physical information and introduces them to T2V models for physics-aware generation. Specifically, it decomposes physical principles into textual physics descriptions, qualitative physics categories, and quantitative physical properties, and designs appropriate conditional injection methods for each type of information. The \textbf{textual physical description} outlines the physical principles to be considered in the scene, the resulting physical phenomena, and their specific visual manifestations. WISA concatenates it with caption before text encoder. \textbf{Qualitative physics categories} indicate the types of physical phenomena that may be involved in the scene. WISA considers 17 common physical phenomena across three major branches of physics (i.e., dynamics, thermodynamics, and optics), such as collision in dynamics, refraction in optics, and melting in thermodynamics.  Considering that different physical phenomena require distinct physical feature, inspired by MoE \cite{riquelme2021scaling} and MoH \cite{jin2024moh}, WISA propose \textbf{M}ixture-of-\textbf{P}hysical-Experts \textbf{A}ttention (\textbf{MoPA}), which assigns expert heads to each physics category, with only the relevant expert heads activated during sampling to handle the corresponding physical phenomena.  \textbf{Quantitative physics properties} represent physical quantities closely related to the physical processes (e.g., density, time, and temperature). WISA encodes these properties as physical embeddings and injects them into the model via AdaLN \cite{peebles2023scalable}.  In addition, WISA employs a Physical Classifier, which is designed to recognize qualitative physics categories, to assist in perceiving physical properties. 

\begin{figure}
  \centering
    \includegraphics[width=1.0\linewidth]{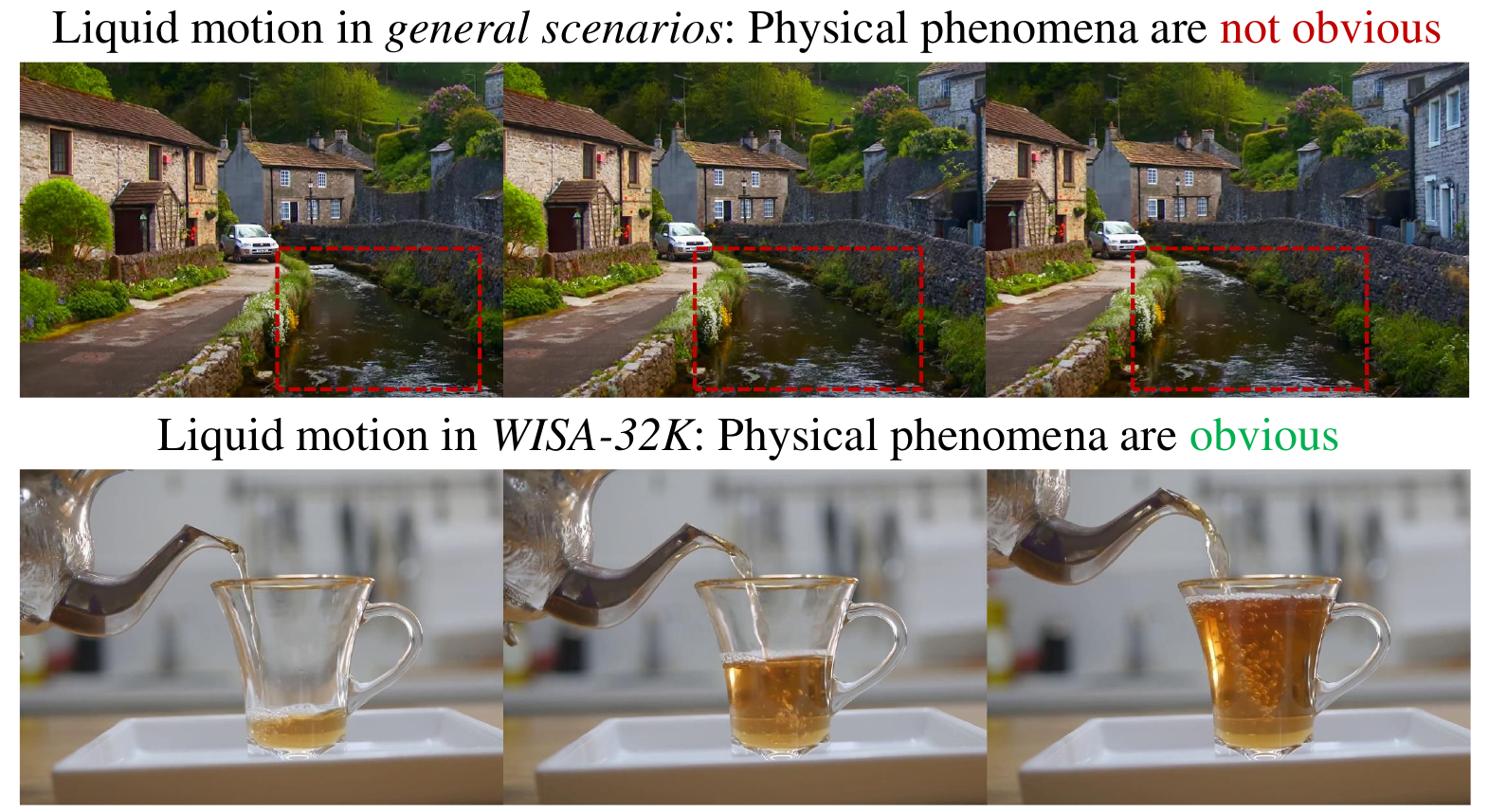}
    \vspace{-15pt}
  \caption{Comparison between general scene videos in Koala-36M and videos with distinct physical phenomena in WISA-32K.}
  \label{fig:data_compare}
  \vspace{-15pt}
\end{figure}

However, extracting above physical information from general scene video in existing datasets \cite{nan2024openvid, wang2024koala} is a suboptimal approach.  Firstly,  general scene videos often feature the interweaving of multiple physical phenomena.  Individual physical phenomena are not prominently visualized, which makes it difficult to accurately extract physical information and establish a precise connection between the physical data and its corresponding visual manifestation.  Secondly, in these datasets, only a few videos distinctly highlight specific physical phenomena as representative examples, while most videos treat physical phenomena as secondary elements.  For instance, in the Figure. \ref{fig:data_compare}, the flow of water is a secondary element. Despite having physical information guidance, the T2V models is unable to perceive the physical principles of fluid motion from this type of data.

To address these challenges, we collect and construct \textbf{WISA-32K}, a dataset containing \textbf{32,000} videos that represent 17 physical phenomena across three major branches of physics as shown in Figure.~\ref{fig:intro}, designed as a data assistant for world simulators. Specifically, based on the previously defined physics categories, we collect videos that clearly exhibit obvious physical phenomena corresponding to each category (e.g., as shown in the lower part of Figure.~\ref{fig:data_compare}). We then apply shot boundary detection, aesthetic quality filtering, and video captioning to the raw videos. Subsequently, we leverage GPT-4o mini to extract and decompose the physical information from the video captions into textual physics descriptions, qualitative physics categories, and quantitative physics properties for WISA.

Our contributions can be summarized as follows:
\begin{itemize}
    \item We propose a physical principle decoupling method, bridging the gap between physical laws and generative modeling. In this method, physical principles are represented as structured physical information, encompassing textual physical descriptions, qualitative physics categories, and quantitative physical properties.
    \item We present the World Simulator Assistant (WISA), which guides T2V models to efficiently learn specific physical phenomena based on structured physical information, through specialized designs such as Mixture-of-Physical-Experts Attention (MoPA) and Physical Classifier.
    \item We manually collect 32,000 video clips that clearly showcase physical phenomena, creating the first large-scale physics video dataset, WISA-32K. It broadly covers common physical phenomena observed in the real world, encompassing 17 types of physical events (e.g., Collision, Melting, and Reflection) across three major branches of physics: Dynamics, Thermodynamics, and Optics.
    \item Quantitative and qualitative experimental results demonstrate WISA and WISA-32K can effectively assist basic T2V models in producing videos that better align with real-world physical laws, while introducing only a 3.5\% increase in parameter count and 5\% inference time.
\end{itemize}

\section{Related Work}
\label{sec:Related_Work}
\subsection{Text-to-Video Generation}
Early text-to-video (T2V) generation research \cite{guo2023animatediff, girdhar2024factorizing, chen2024videocrafter2, wang2024qihoo, feng2024fancyvideo, wang2023modelscope, blattmann2023stable} primarily extend image generation models \cite{blattmann2023align, lipman2022flow, li2024hunyuan, podell2023sdxl, liu2023bridge, ma2024hico} with temporal capabilities to enable video generation. These methods often suffered from limited realism and restricted motion dynamics. The powerful 3D spatio-temporal modeling and scalability of Diffusion Transformers \cite{peebles2023scalable, flux2024} have greatly advanced the development of visual generation models. Enabled by Diffusion Transformers, a series of recent T2V works (including OpenSora \cite{opensora}, Cosmos \cite{agarwal2025cosmos}, Sora \cite{openai2024videogeneration}, CogVideoX \cite{yang2024cogvideox}, HunyuanVideo[x], Kling \cite{kling2024videogeneration}, Wan2.1 \cite{wan2.1}, and Step-Video-T2V \cite{ma2025step}) significantly improve the realism and motion quality of video generation by scaling up model parameters and training data. These works are widely considered as a promising pathway towards building a World Simulator. However, they still struggle to generate videos that fully comply with real-world physical laws as they essentially fit the data distribution \cite{kang2024far} from general-scene datasets such as Koala-36M \cite{wang2024koala} and OpenVid \cite{nan2024openvid}, where physical laws are not explicitly reflected and physical phenomena are not prominently presented (e.g., in the upper part of Figure. ~\ref{fig:data_compare}). In contrast, our carefully curated WISA-32K dataset prioritizes the explicit presentation of typical physical phenomena as the primary criterion for video collection as presented in Figure. ~\ref{fig:intro}. And it provides detailed and structured physical information annotations, making it a valuable data assistant for enhancing the physical consistency of video generation.

\subsection{Physical-aware Video Generation}
Recently, researchers \cite{motamed2025generative,bansal2024videophy,meng2024towards,le2023differentiable,aira2024motioncraft,liu2024physgen,xue2024phyt2v, wang2024lavie} have increasingly focused to improving and evaluating the physical consistency of generated videos. On the one hand, Videophy \cite{bansal2024videophy} and PhyGenBench \cite{meng2024towards} build test samples that reflect various physical laws, and they evaluate how well generated videos follow real-world physical laws by either training physics classification models with manual annotations or using question-answering methods based on Vision-Language models \cite{wang2024internvideo2}. On the other hand, DANO \cite{le2023differentiable}, MotionCraft \cite{aira2024motioncraft}, and PhysGen \cite{liu2024physgen} parse objects from images and estimate their rigid motion in a differentiable manner by considering physical properties such as mass, inertia, friction, and rotation. Based on these estimations, they animate the images into videos. However, these methods are restricted to fixed physical categories (e.g., rigid motion) and static scenarios that involve only object motion, which hinders their generalizability. PhyT2V \cite{xue2024phyt2v} leverages large language models and vision-language models to extract physical inconsistency information from generated videos. Based on the extracted physical feedback, it iteratively refines the textual description over multiple rounds, improving video generation quality. Although this approach offers generality, it introduces significant inference overhead and fails to enhance the generative model’s ability to encode physical knowledge. In this paper, WISA incorporates structured physical information into the generative model, enhancing its physical perception and enabling it to handle various physical phenomena more effectively.
\section{WISA-32K}

\subsection{Data Collection and Annotation}
\label{define}

\begin{figure*}
  \centering
    \includegraphics[width=1.0\linewidth]{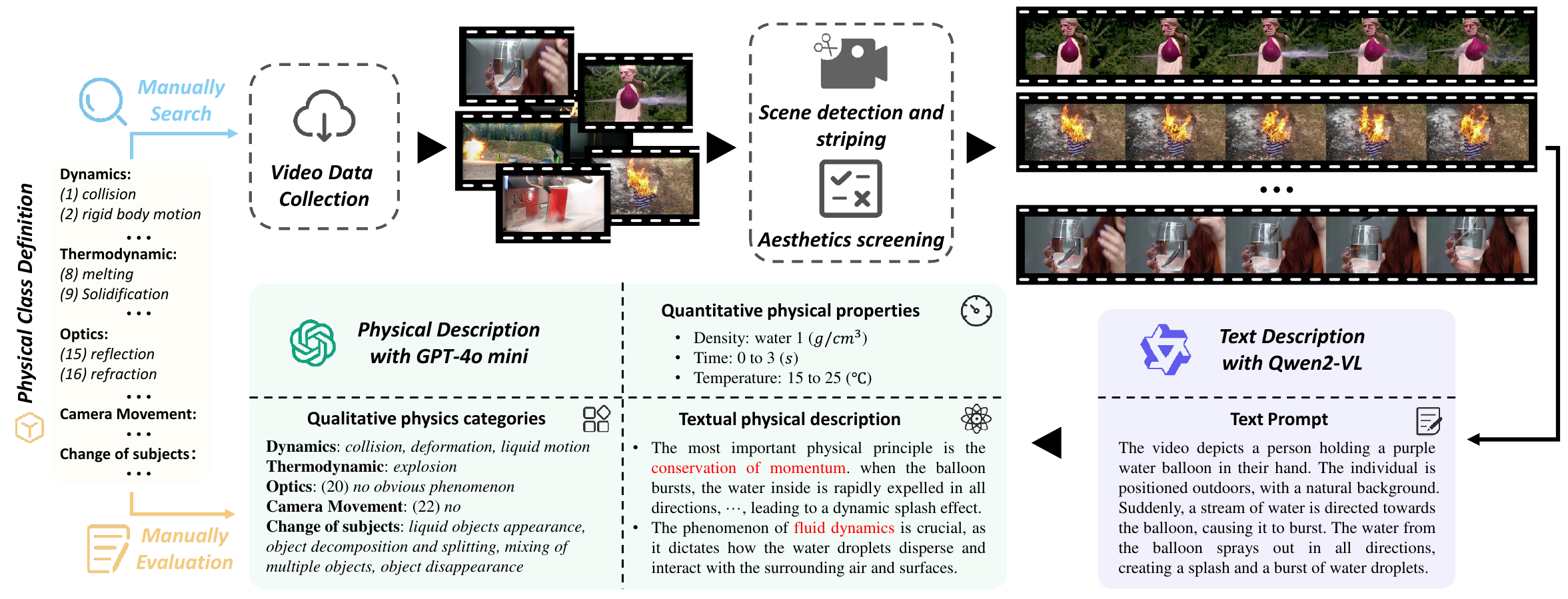}
    \vspace{-15pt}
  \caption{Pipeline of WISA-32K. We first define 17 common physical phenomena and, based on this, manually collect 32,000 video samples that clearly illustrate these phenomena. Then, we perform shot detection and aesthetic filtering on the raw videos. Text description are extracted using Qwen2-VL, and detailed physical annotations are generated with GPT-4o mini.}
  \label{fig:data_pipe}
  \vspace{-12pt}
\end{figure*}

\textbf{Physical Laws Definition:}
We select three fundamental categories of physics that are universally relevant in life: \textit{Dynamics}, \textit{Thermodynamics}, and \textit{Optics}. Seventeen physical phenomena associated with these categories are then considered in WISA-32K.

\textit{\textbf{Dynamics}}: We consider six common dynamic phenomena encountered in daily situations: \textit{Collision}, \textit{Rigid Body Motion}, \textit{Elastic Motion}, \textit{Liquid Motion}, \textit{Gas Motion}, and \textit{Deformation}. For instance, the swinging of a pendulum serves as an example of \textit{Rigid Body Motion}.

\textit{\textbf{Thermodynamics}}: We select six common thermodynamic phenomena observed in typical life scenarios: \textit{Melting}, \textit{Solidification}, \textit{Vaporization}, \textit{Liquefaction}, \textit{Explosion}, and \textit{Combustion}. For example, a time-lapse of melting ice cream illustrates the \textit{Melting} phenomenon.

\textit{\textbf{Optics}}: We define five common optical phenomena: \textit{Reflection}, \textit{Refraction}, \textit{Scattering}, \textit{Interference and Diffraction}, and \textit{Unnatural Light Sources}. For example, a video showing the reflection on a lake illustrates the \textit{Reflection}.

Based on the 17 physical phenomena outlined above, we manually collected 32,000 video samples, intentionally excluding videos with text. Additionally, we did not consider certain physical phenomena (e.g., sublimation, condensation) due to their infrequent occurrence in life and the challenges associated with collecting data for these phenomena.

\noindent \textbf{Pre-processing and Caption:} The video data is manually collected, ensuring the exclusion of videos containing text or low-quality content. Consequently, only simple pre-processing techniques are applied.
We use PySceneDetect \cite{PySceneDetect} to split the raw videos into individual scene clips, followed by filtering based on aesthetic scores. Then, we utilize Qwen2-VL \cite{wang2024qwen2} to generate video captions using the following prompt: \{\textit{Please describe the content of this video in as much detail as possible, including the objects, scenery, animals, and camera movements within the video.}\} The caption length is limited to 256 tokens.

\subsection{Physical Information Decompose}
We believe that simple video captions are not sufficient to clearly represent the physical information and related physical phenomena in a video. As shown in the Figure. \ref{fig:data_pipe}, we further constructed structured physical annotations to analyze the physical information from multiple dimensions. Specifically, we decompose the physical information into: \textit{textual physical descriptions}, \textit{qualitative physics categories}, and \textit{quantitative physical properties}.

\textbf{Textual physical descriptions}: Provide a detailed explanation of the physical principles to be considered and the resulting intuitive physical phenomena, while supplementing the missing physical information in the prompt.

\textbf{Qualitative physics categories}: Based on the physical laws defined in Sec.~\ref{define}, we annotate the physical phenomena present in each video and identify which of the 17 physical phenomena are involved. Three categories of anomalies (i.e., \textit{No obvious dynamic phenomenon}, \textit{No obvious thermodynamic phenomenon}, and \textit{No obvious optical phenomenon}) are introduced to account for scenarios that do not involve dynamics, thermodynamics, or optical phenomena. Furthermore, nine categories of visual phenomena are introduced, two of which pertain to whether the shot exhibits motion, while the remaining seven correspond to changes in the state of moving entities (i.e., \textit{Object decomposition and splitting}, \textit{Mixing of multiple objects} ... The detailed explanation please refer to \textcolor{blue}{\textbf{Supplementary Material \ref{categories_def}}}). There are a total of 29 qualitative physics categories.

\textbf{Quantitative physical properties}: Three physical attributes related to multiple physical phenomena are annotated, namely the density of primary motion physics, the time range during which the physical phenomenon occurs, and the temperature range during which the physical phenomenon occurs.

Due to the significant computational overhead and cost associated with video multi-modal models, the annotation of the above physical information is carried out using GPT-4o mini based on caption. Specifically, we conduct five rounds of annotation to label qualitative physical phenomenon categories (i.e., dynamics, thermodynamics, optics, motion, the state of objects), and three rounds to annotate quantitative physical attributes (i.e., \textit{Density}, \textit{Time} and \textit{Temperature}). Detailed annotation prompts and example are provided in the \textcolor{blue}{\textbf{Supplementary Material \ref{example_annotation} and \ref{annotation_prompts}}}. 

We sample 100 examples and manually evaluate the multi-modal annotation scheme and the caption-based annotation scheme using GPT-4o mini. Thanks to the accurate captions provided by Qwen2-VL, the caption-based annotation scheme achieves performance only slightly lower than the multi-modal scheme (76\% vs. 78\%). However, the caption-based annotation method offers substantial cost advantages (approximately 2k vs. 10k tokens per sample) and provides greater convenience for users when only textual descriptions are available. More analysis of WISA-32K please refer to the \textcolor{blue}{\textbf{Supplementary Material \ref{property_analysis} and \ref{word_cloud}}}.

\section{Method}

\begin{figure*}
  \centering
    \includegraphics[width=1.0\linewidth]{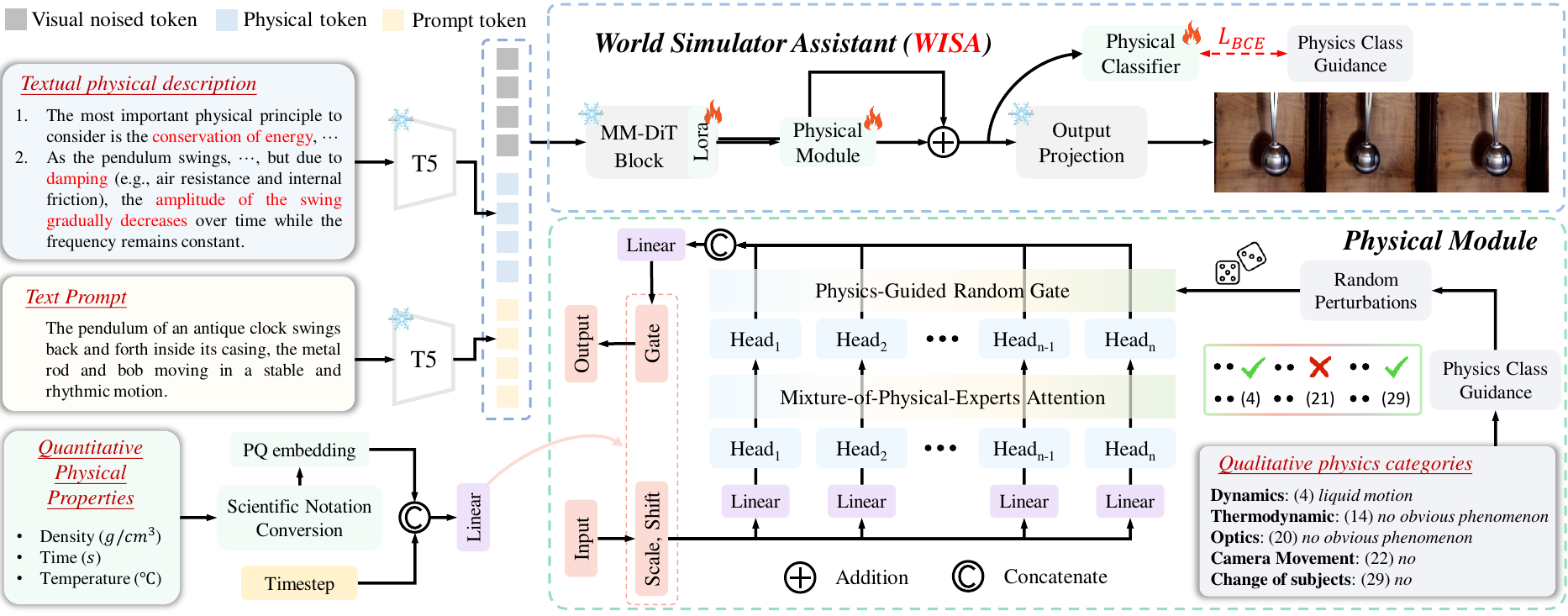}
  \caption{Overview of the proposed WISA. WISA introduces the Physical Module and Physical Classifier, which leverage structured physical annotations to guide and assist T2V models in generating physics-aware videos. Specifically, for qualitative physical categories, WISA constructs a Mixture-of-Physical-Experts Attention within the Physical Module, where each attention head corresponds to a specific physical phenomenon. The relevant physical expert is activated by the input qualitative physical category. The Physical Classifier predicts the physical categories relevant to the video and is supervised by inputted categories to understand abstract physical principles.}
  \label{fig:method}
\end{figure*}

\subsection{Overview}
Given textual physical descriptions, qualitative physical categories, and quantitative physical properties, we design the WISA framework to efficiently incorporate these conditions into existing T2V models (i.e., CogVideoX \cite{yang2024cogvideox}). To facilitate the learning of physical knowledge while preserving the model's original capabilities with limited video data, we design three distinct condition injection methods tailored to each of the three categories of physical information, as illustrated in Figure. \ref{fig:method}. Specifically, for the textual physical descriptions, we concatenate them with the video caption and leverage the generative model's inherent semantic understanding to generate visual phenomena described in text (Such as "amplitude of the swing gradually decreases over time" in Figure. \ref{fig:method}).
For qualitative and quantitative physical conditions, WISA introduces the Physical Module. In this module, we propose a Mixture-of-Physical-Experts Attention (MoPA), which assigns expert heads to each physics category to model category-specific features. Quantitative physical quantities are encoded as physical embeddings and then integrated into the denoising feature within the module using AdaLN. Additionally, we introduce a qualitative Physical Classifier to help the model understand the physical conditions.  Due to the significant computational and parameter cost introduced by MoPA, only one physical module is inserted after all the Diffusion transformer blocks to accelerate training and reduce the overall burden. Detailed explanations and elaborations of the Physical Module and Physical Classifier are provided in Sec. \ref{method:PM} and Sec. \ref{method:PC}.

\subsection{Physical Module}
\label{method:PM}
Most videos from real-world scenes involve the coupling of multiple physical phenomena. Even when decomposed into distinct physical categories in WISA-32K, it remains challenging for T2V models to comprehend the abstract qualitative physical categories and accurately model specific types of physical phenomena. To address this challenge, we propose a Mixture-of-Physical-Experts Attention within the Physical Module. Inspired by MoH \cite{jin2024moh}, this mechanism assigns each head in the multi-head self-attention to a specific class of physical phenomena and activates the output of the relevant head only when the corresponding phenomenon is present. This approach treats each head as an expert in its domain, enabling it to independently model the properties of a particular physical phenomenon.

Specifically, qualitative physical categories are encoded as $P_c \in \mathbb{R}^{C}$, where $C$ denotes the number of defined physical phenomena (i.e., 29). Here, $P_c^i = 0$ indicates that the corresponding category is not activated, and $P_c^i = 1$ indicates that the corresponding category is activated, with $i$ being the category index. 
Physical categories cannot be absolutely correct and may contain noise, such as incorrect activations or suppressions.  To mitigate the impact of these noises on training, we employ a random perturbation operation,  where the positions with $P_c^i = 1$ are set to 0.1 and the positions, and the positions with $P_c^i = 0$ are set to 1.0 with a certain probability (i.e., 0.2), resulting $\hat{P_c}$. 
After the multi-head self-attention operation, the denoising feature $F_h \in \mathbb{R}^{N \times d \times h}$ (where $h$ presents the number of head and $h = C$, and $d$ denotes head dimension) will interact with $\hat{P_c}$ to activate and suppress the experts corresponding to different physical phenomena. The feature dimension is then restored through concatenation and a linear layer. The mathematical representation of this process is as follows:
\begin{equation}
\begin{array}{l}
  \hat{P_c} = \mathrm{Random}(P_c), \, F_h =\mathrm{MHSA}(F), \\
  F_o = \mathrm{Linear}(\mathrm{Reshape}(F_h \odot \hat{P_c})) \\
  \label{eq:pm}
\end{array}
\end{equation}
where $\mathrm{Random}$ denotes random perturbations operation, $\mathrm{MHSA}$ represents multi-head self-attention, and $\odot$ denotes  element-wise multiplication. 

Due to the large variations in the time and temperature spans of different physical phenomena, we first represent the temperature and time in the quantitative information using scientific notation, with coefficients and exponents. These values are mapped through a linear layer, concatenated with the timestep embedding, and injected by AdaLN.

Generative models often consist of multiple transformer blocks with large feature dimensions, inserting the Physical Module after every block would lead to an explosion in both parameters and computational complexity. Additionally, it could result in a loss of the model’s inherent capabilities and cause slower convergence of the shallow Physical Module. Therefore, we insert the Physical Module only after the final transformer block, achieving efficient physical information guidance while mitigating the aforementioned issues.

\subsection{Physical Classifier}
\label{method:PC}
To guide the generative model in understanding abstract physical categories and modeling physical properties, we introduce a Physical Classifier after the Physical Module to predict qualitative physical categories.
Multiple physical phenomena may be coupled in a video, we use a multi-label binary cross-entropy (BCE) loss for supervision.
\begin{equation}
    L_{pc} = \sum_{i=1}^C(P_c^i\mathrm{log}(f_c^i) + (1 - P_c^i)\mathrm{log}(1 - f_c^i)),
  \label{eq:loss}
\end{equation}
where $C$ is the number of physical categories, and $f_c \in \mathbb{R}^{C}$ represents the predicted probabilities, which are obtained by passing the denoising feature through the Physical Classifier and the sigmoid function.

To balance the introduced classification loss $L_{pc}$ and the diffusion loss $L_{diffusion}$, we adopt the following loss function to optimize the physics-aware generative model.
\begin{equation}
    L = L_{diffusion} + \lambda L_{pc} / (1 + L_{pc}.\mathrm{detach}), 
  \label{eq:total_loss}
\end{equation}
where $\lambda$ is balance coefficient.
\section{Experiments}
\subsection{Setup}
\paragraph{Training Setting:} We select the current representative open-source T2V model, CogVideoX-5B, as the base T2V model to validate the effectiveness of WISA. More training detail please refer to \textcolor{blue}{\textbf{Supplementary Material \ref{traing_detail}}}.

\paragraph{Evaluation:} We select VideoCon-Physics from Videophy \cite{bansal2024videophy} to evaluate the physical law consistency (PC) and semantic coherence (SA) of the generated videos. We use 160 carefully crafted prompts from PhyGenBench \cite{meng2024towards} and 344 prompts from Videophy, designed to reflect various physical principles, for testing. We consider PC and SA return values greater than or equal to 0.5 as PC = 1 and SA = 1, and values less than 0.5 as PC = 0 and SA = 0.

\begin{figure*}
  \centering
    \includegraphics[width=1.0\linewidth]{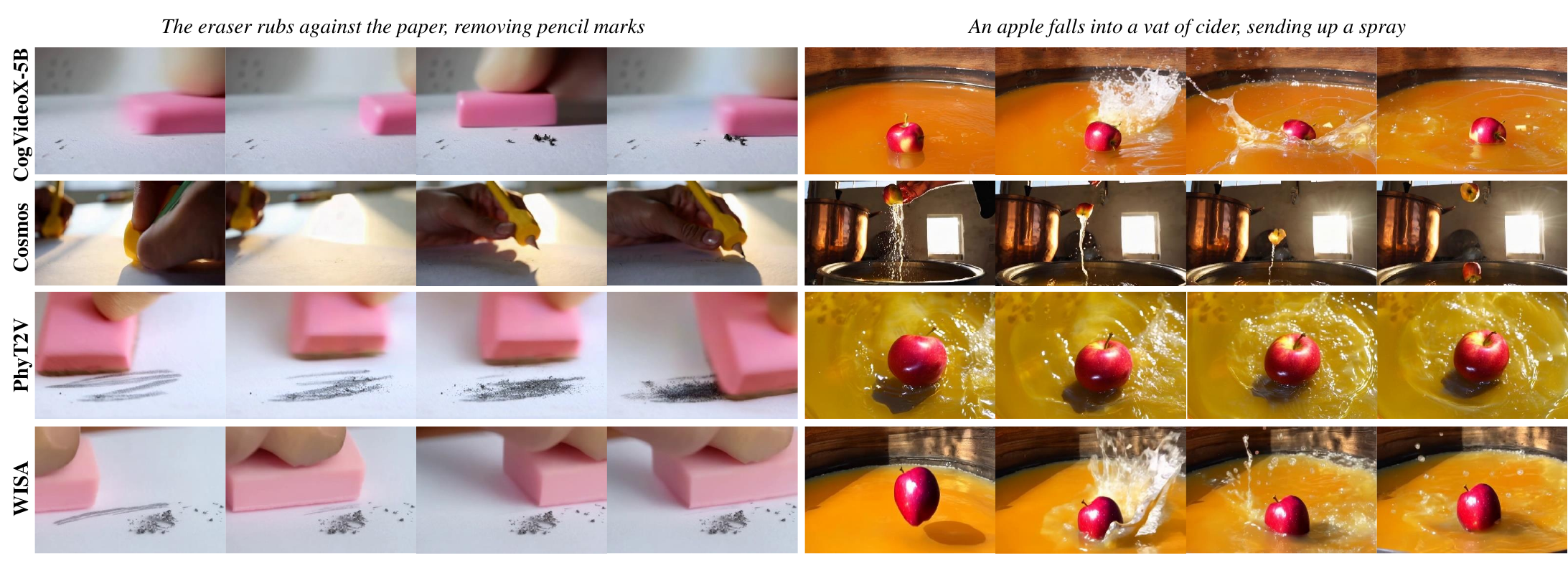}
    \vspace{-25pt}
  \caption{Qualitative comparison between WISA and existing T2V methods. WISA exhibit better alignment with real-world physical laws.}
  
  \label{fig:compare_visual}
\end{figure*}

\begin{figure*}
  \centering
    \includegraphics[width=1.0\linewidth]{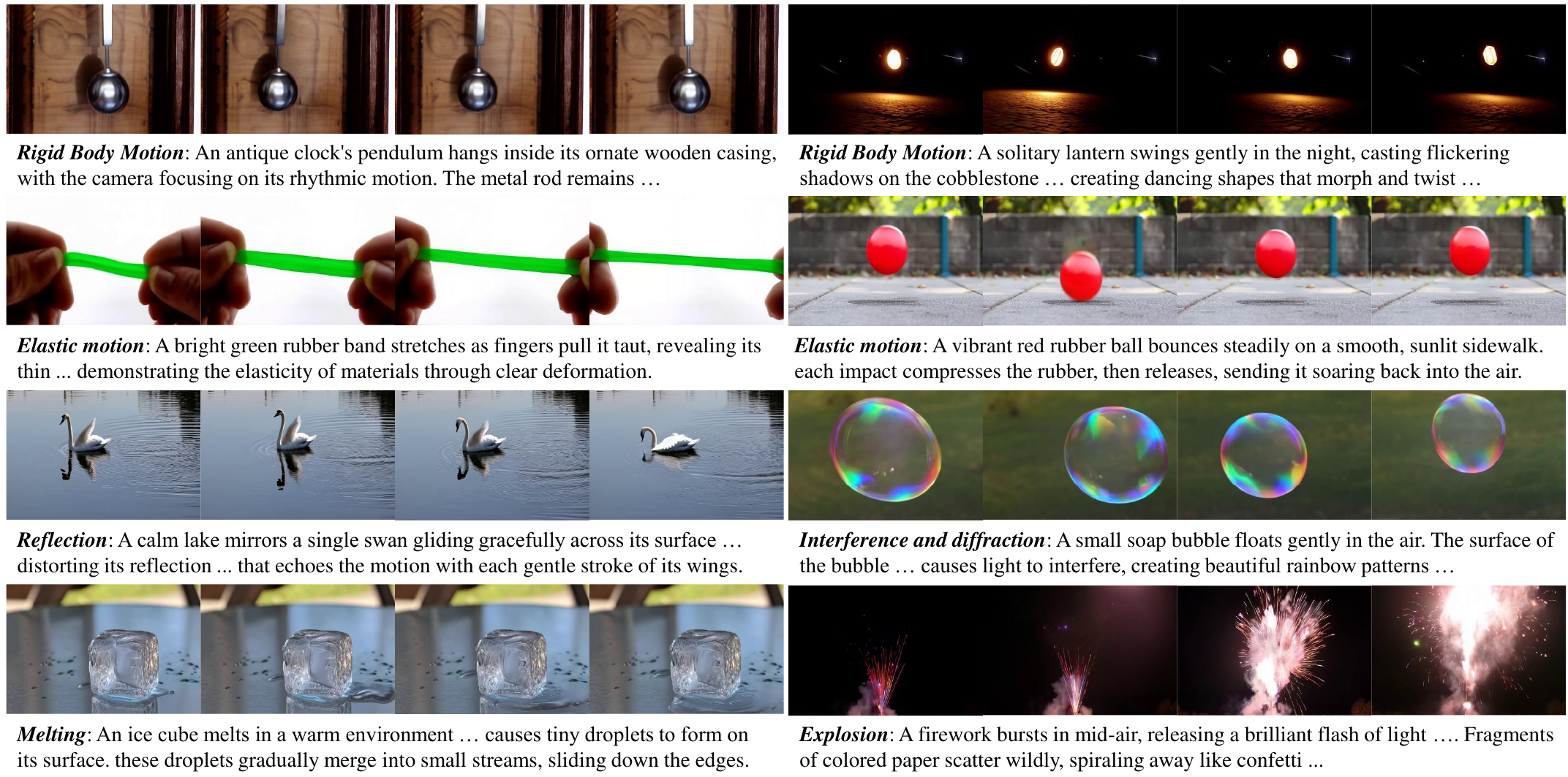}
    \vspace{-20pt}
  \caption{More samples generated by WISA, covering additional physical phenomena.}
  \label{fig:our_visual}
  \vspace{-10pt}
\end{figure*}

\subsection{Quantitative comparison}
\begin{table}
\footnotesize
\caption{Quantitative evaluation using VideoCon-Physics conducted on the Videophy and PhyGenBench prompt lists. The best and second performing metrics are highlighted in \textbf{bold} and \underline{underline} respectively. $^*$ denotes the scores reproduced by us.}
\setlength{\tabcolsep}{0.95mm}
\begin{tabular}{c|c|cc|cc}
    \toprule[1.0pt]
    \multirow{2}{*}{Method}  & Inference & \multicolumn{2}{c|}{VideoPhy \cite{bansal2024videophy}} & \multicolumn{2}{c}{PhyGenBench \cite{meng2024towards}} \\
    \cmidrule{3-6}
     & Time (s) & \multicolumn{1}{c}{SA ($\uparrow$)} & \multicolumn{1}{c|}{PC ($\uparrow$)} & \multicolumn{1}{c}{SA ($\uparrow$)} & \multicolumn{1}{c}{PC ($\uparrow$)} \\
     \midrule
      VideoCrafter2 \cite{chen2024videocrafter2} & - & 0.47 & \underline{0.36} & - & -  \\
    HunyuanVideo \cite{li2024hunyuan} & - & 0.46 & 0.28 & - & -  \\
     CogvideoX-5B$^*$ \cite{yang2024cogvideox} & 210 & 0.60 & 0.33 & 0.39 & 0.41  \\
     Cosmos$^*$ \cite{agarwal2025cosmos} & 600 & 0.57 & 0.18 & \textbf{0.43} & 0.14  \\
      \gr{PhyT2V (Round 4)} \cite{xue2024phyt2v} & \gr{1800} & \gr{0.59} & \gr{0.42} & \gr{0.38} & \gr{\underline{0.42}} \\
     PhyT2V$^*$ (Round 4) \cite{xue2024phyt2v} & 1800 & \underline{0.61} & 0.37 & - & -  \\
     \midrule
     WISA & 220 & \textbf{0.67} & \textbf{0.38} & \underline{0.40} & \textbf{0.43}  \\
     \bottomrule[1.0pt]
\end{tabular}
\vspace{-10pt}

\label{sota}
\end{table}

We select four general text-to-video generation models (i.e., VideoCrafter2, HunyuanVideo, CogVideoX-5B and Cosmos-Diffusion-7B) and PhyT2V, a method specifically designed to enhance physical properties, for quantitative comparison, as shown in Table. \ref{sota}.

\textbf{VideoPhy}:
WISA achieves state-of-the-art performance on both SA and PC metrics, while maintaining high efficiency. Compared to the baseline (CogVideoX-5B), WISA improves SA and PC scores by 0.07 and 0.05, respectively, demonstrating that our proposed method significantly enhances the realism of generated videos. PhyT2V improves its performance by iteratively analyzing physical errors in generated video captions and adjusting the input prompts based on feedback from VideoCon-Physics scores. However, its cumbersome pipeline, which involves multiple rounds of Tarsier-34B \cite{wang2024tarsier} inference for video generation, introduces extremely long inference time—approximately 9 times longer than the original generation model (CogVideoX-5B). Cosmos exhibits poor performance due to the disordered physical processes and inconsistent temporal sequences.

\textbf{PhyGenBench}: We also evaluate our method on the prompts from PhyGenBench, achieving SOTA results, which demonstrates the generalizability of WISA.

\subsection{Qualitative comparison}
We further provide a qualitative comparison with existing methods to demonstrate the advantages of WISA. As shown in the Figure. \ref{fig:compare_visual}, for the example of erasing pencil marks with an eraser, WISA generates a video where the pencil marks are cleanly removed as the eraser passes over them. In contrast, CogVideoX-5B fails to generate any pencil marks, PhyT2V makes the pencil marks even darker after erasing, and Cosmos does not show the erasing process at all. In the example on the right, WISA successfully simulates the process of an apple falling into water: the water surface remains calm before the apple enters, splashes form as the apple impacts the water, and the apple experiences buoyant force after submersion. However, CogVideoX-5B generates chaotic water and apple movements, PhyT2V omits the falling process, and Cosmos mistakenly generates two apples at the end. Additional videos generated by WISA, demonstrating various physical phenomena, are also presented in the Figure. \ref{fig:our_visual}. All the videos mentioned above are available in \href{https://wisav1.github.io/WISA/}{Project Page}.

\subsection{Ablation Study}
We conduct ablation studies on VideoPhy using VideoCon-Physics to verify the effectiveness of key components in our method, as shown in the Table. \ref{tab:ablation}. As expected, removing the Physical Module leads to a performance drop, due to the lack of qualitative and quantitative physical information guidance. Similarly, the Physical Classifier helps the generative model perceive and model physical properties, which benefits both semantic consistency and physical law consistency.
Moreover, the training data of the evaluation model VideoCon-Physics \cite{bansal2024videophy} comes from samples generated by nine T2V models, which leads to a distribution shift compared to the real-world videos in WISA-32K. As a result, solely relying on LoRA brings only limited improvements. Furthermore, we explore the role of clearly-defined physical phenomena data and general scene data in enhancing physical perception. We sample 32,000 videos from Koala-36M, label the physical information, and train WISA, which results in limited improvement. This showcases that videos with clearly physical phenomena in WISA-32K are highly beneficial for modeling physical properties.

\subsection{Human Evalution}
\begin{figure}[t]
  \centering
    \includegraphics[width=1.0\linewidth]{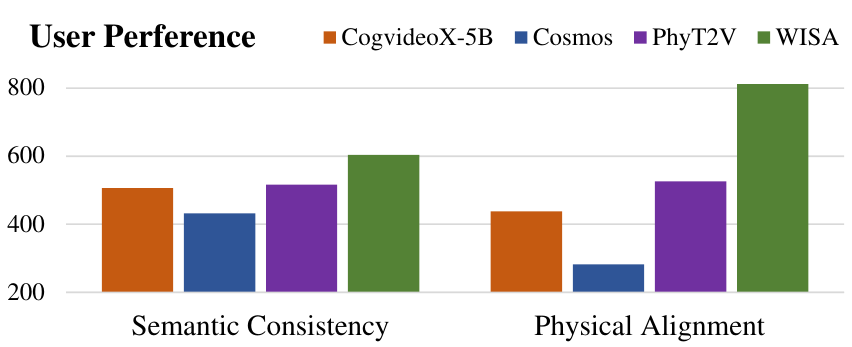}
    \vspace{-20pt}
  \caption{Human evaluation on VideoPhy prompts.}
  \label{fig:human_eval}
  \vspace{-20pt}
\end{figure}

The physical consistency of generated videos is abstract and difficult to quantify directly. Therefore, we conduct a human evaluation to assess the effectiveness of WISA. Specifically, we selected three representative models for comparison. The evaluation considered two aspects: semantic consistency and physical alignment. Each candidate model is ranked in both aspects, receiving a score based on its ranking: 3 points for first place, 2 points for second, and 0 points for last. The results, shown in Figure. \ref{fig:human_eval}, demonstrate that WISA achieves a significant advantage in physical alignment, while also maintaining strong semantic consistency.

\begin{table}
\centering
\footnotesize
\caption{Ablation study on the key components of WISA.}
\vspace{-10pt}
\setlength{\tabcolsep}{3.0mm}
\begin{tabular}{c|c|cc}
    \toprule[1.0pt]
    & Setting & SA ($\uparrow$) & PC ($\uparrow$)  \\
     \midrule
    \multirow{4}{*}{Structure} & Baseline & 0.60 & 0.33  \\
     & only LoRA & 0.64 & 0.34  \\
      & w/o Physical Module & 0.64 & 0.33  \\
      & w/o Physical Classifier & 0.66 & 0.36 \\
    \midrule
   \multirow{2}{*}{Data} & 32K sample from Koala-36M & 0.62 & 0.33 \\
      & WISA-32K & 0.67 & 0.38  \\
     \bottomrule[1.0pt]
\end{tabular}
\vspace{-10pt}
\label{tab:ablation}
\end{table}

\subsection{Attention Map Analysis}
We further conduct a visual analysis of the Mixture-of-Physical-Experts attention maps, aiming to investigate whether different physical experts focus on the regions corresponding to distinct physical phenomena. As shown in the Figure. \ref{fig:data_correct}, the rigid body motion expert perfectly focuses on the swing region, while the non-dynamics expert attends to the static background with no apparent motion. This demonstrates that the MoPA effectively models and captures the corresponding physical attributes.

\begin{figure}[t]
  \centering
    \includegraphics[width=1.0\linewidth]{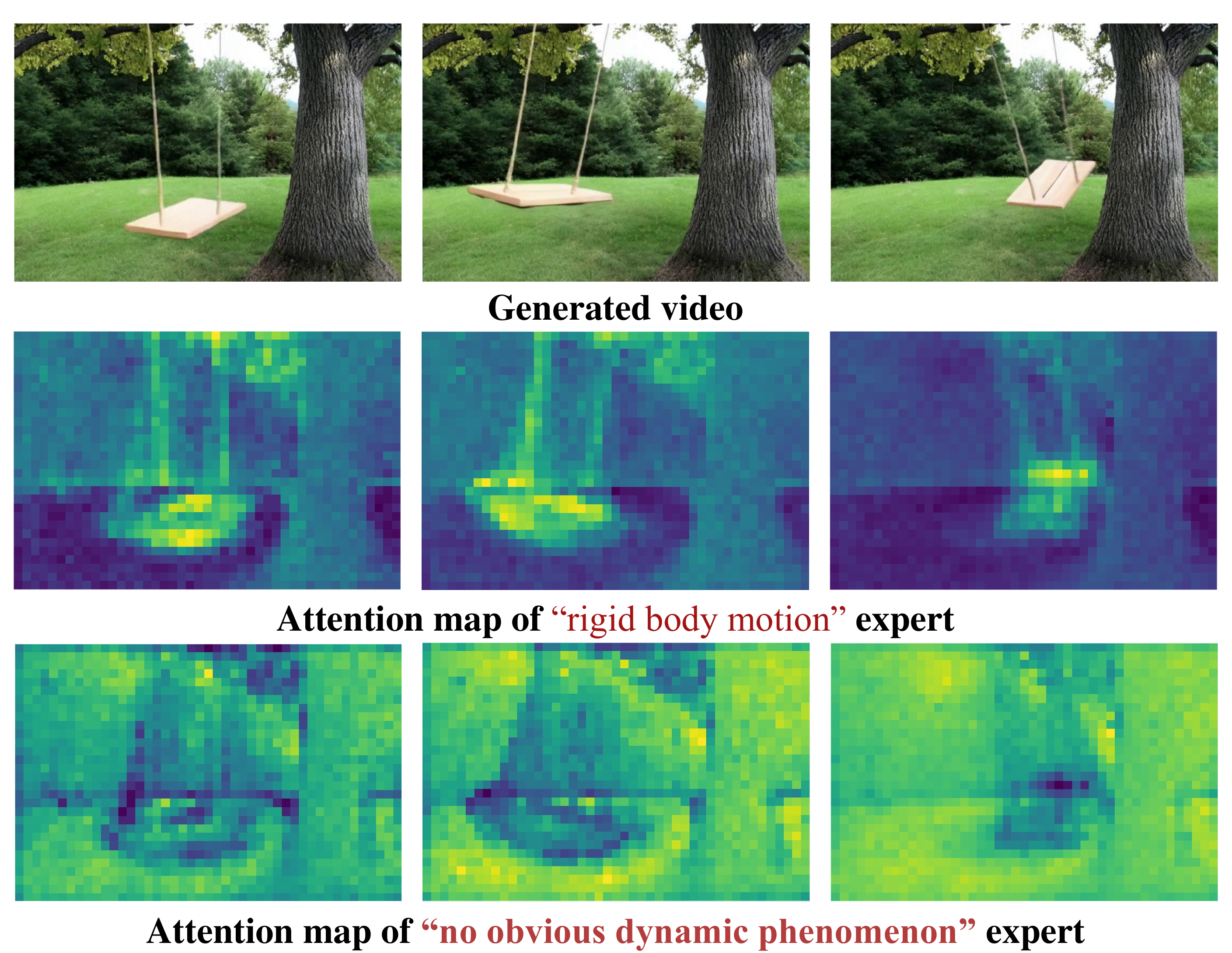}
    \vspace{-15pt}
  \caption{Attention maps of different physical experts.}
  \label{fig:data_correct}
  \vspace{-18pt}
\end{figure}
\section{Limitation}
Although our approach significantly improves the ability of existing T2V models to generate videos that align with real-world physical laws, it still has the following limitations: 1) Limited physical categories: We collect 32,000 videos in WISA-32K, covering 17 types of physical phenomena. However, due to constraints in time and manpower, the dataset does not include all physical phenomena encountered in real world, such as corrosion or vacuum environments. 2) Limited physical information guidance: WISA primarily provides high-level semantic guidance and lacks detailed constraints at the physical mechanism level (e.g., energy conservation, Newton's laws). However, introducing more detailed physical principle constraints currently requires modeling object motion based on image or 3D information, which suffers from poor generalization and can only handle limited categories and scenarios. How to incorporate physical principle constraints into text-to-video generation while maintaining generalization remains an area worth further research. 3) Failure case: Due to the limited data and parameter, WISA cannot generate videos that perfectly align with physical principles in all scenarios. 

\section{Conclusion} 
In this paper, we present WISA framework, which decomposes physical principles into structured physical information, including textual physical descriptions, qualitative physical categories, and quantitative physical properties. To help T2V models learn these physical aspects effectively, WISA incorporates two key components: the Mixture-of-Physical-Experts Attention and the Physical Classifier. Building on this, we construct WISA-32K, a dataset containing 32,000 video clips that cover 17 physical phenomena across three fundamental categories of physics, providing a high-quality data foundation. Experimental results show that WISA and WISA-32K can effectively help producing videos that better align with real-world physical laws, while the additional computational overhead is under 5\%. We hope that WISA can provide valuable insights to the research on building powerful world simulators.

{
    \small
    \bibliographystyle{ieeenat_fullname}
    \bibliography{main}
}
\clearpage 
\appendix

\section{Training Detail}
\label{traing_detail}
We select the current representative open-source T2V model, CogVideoX-5B, as the base T2V model to validate the effectiveness of the proposed WISA. We train WISA on our constructed WISA-32K for 8,000 steps, using a learning rate of 2e-5 and a batch size of 8. The video resolution is set to 480x720, with 49 frames. The LoRA rank is set to 128, and the LoRA alpha is set to 16. During training, only the physical module, physical classifier, and parameters in LoRA are updated, with a total of 187M learnable parameters. The experiments are conducted on 8 A100 GPUs, each with 80GB of memory.

\section{The Definition of Physical Categories}
\label{categories_def}
We define a total of 29 qualitative physical categories, organized into 5 major classes. The physical categories within each class, along with their corresponding category IDs, are listed as follows:

\textbf{Dynamics}: \textit{1. Collision}, \textit{2. Rigid Body Motion}, \textit{3. Elastic Motion}, \textit{4. Liquid Motion}, \textit{5. Gas Motion}, \textit{6. Deformation}, and \textit{7. No obvious dynamic phenomenon}

\textbf{Thermodynamics}: \textit{8. Melting}, \textit{9. Solidification}, \textit{10. Vaporization}, \textit{11. Liquefaction}, \textit{12. Explosion},  \textit{13. Combustion} and \textit{14. No obvious thermodynamic phenomenon}

\textbf{Optics}: \textit{15. Reflection}, \textit{16. Refraction}, \textit{17. Scattering}, \textit{18. Interference and Diffraction}, \textit{19. Unnatural Light Sources}, and \textit{20. No obvious optical phenomenon}

\textbf{Camera motion}: \textit{21. Yes}, \textit{22. No}

\textbf{The state of object}: \textit{23. Liquids Objects Appearance}, \textit{24. Solid Objects Appearance}, \textit{25. Gas Objects Appearance}, \textit{26. Object decomposition and splitting}, \textit{27. Mixing of Multiple Objects}, \textit{28. Object Disappearance} and \textit{29. No Change}

Specifically, \textit{Liquids objects appearance}: new liquids appear from the camera over time and due to external forces, such as water squeezed out of a towel. \textit{Solid objects appearance}: new solids appear from the camera over time and due to external forces, such as Chemical reaction produces precipitate or car drives in from outside the camera. \textit{Gas objects appearance}: new gas appear from the camera over time and due to external forces. \textit{Object decomposition and splitting}: Over time and under the action of external forces, an object is broken into multiple sub-parts: such as fruits and vegetables being cut in half. \textit{Mixing of multiple objects}: Over time and with the action of external forces, two objects of the same state mix together, such as two solutions mixing. \textit{Object disappearance}: As time passes and external forces act, objects disappear from the camera. \textit{No change}: No change in the state of the object

\begin{figure}[h]
  \centering
    \includegraphics[width=1.0\linewidth]{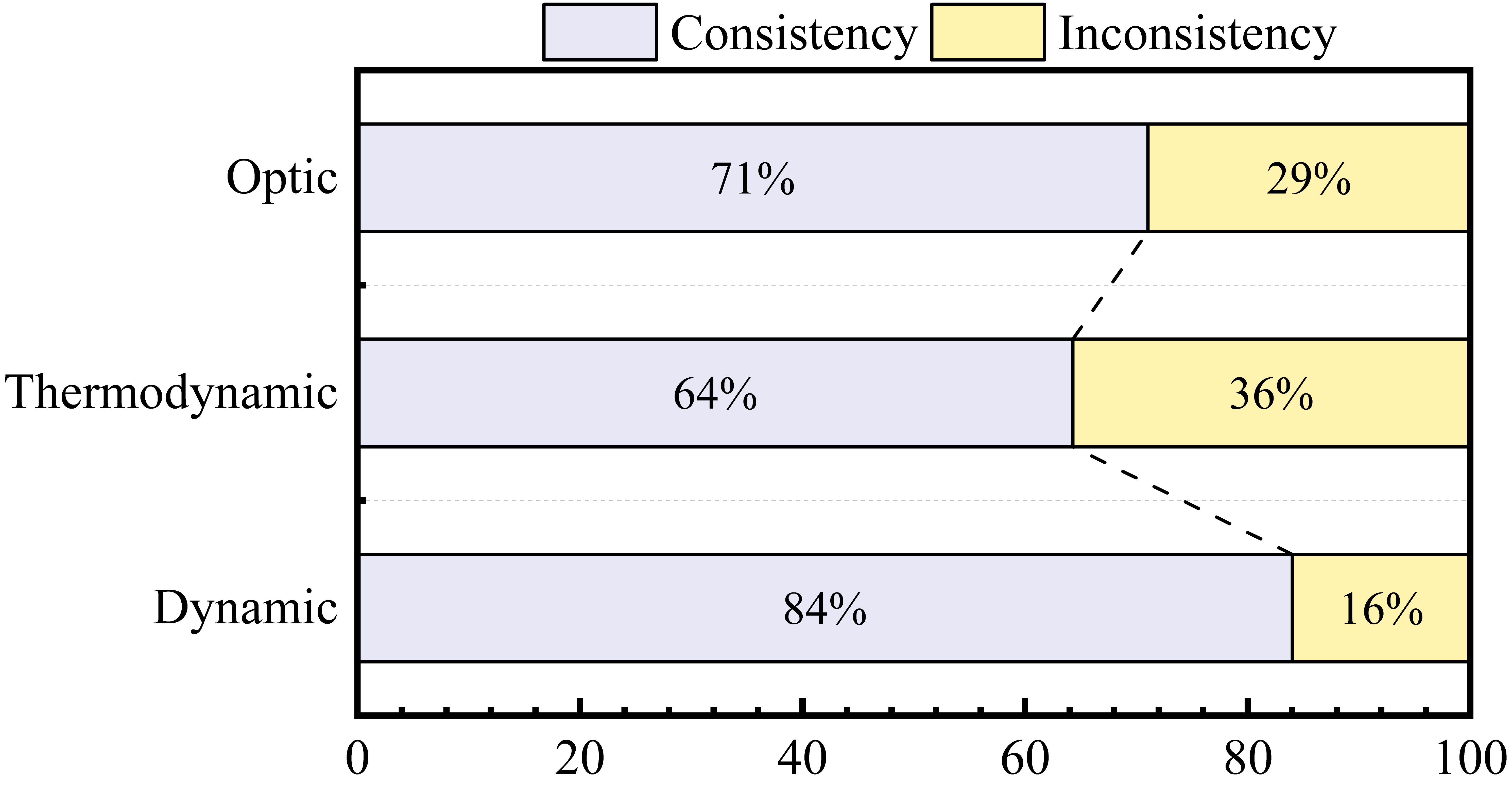}
    \vspace{-20pt}
  \caption{Accuracy of qualitative physical category annotations.}
  \label{fig:data_correct}
  \vspace{-10pt}
\end{figure}

\section{Dataset Property Analysis}
\label{property_analysis}
We visualize the distribution of different physics categories and video frame counts in WISA-32K, as shown in the paper Figure. 1.
Dynamics frequently occurs in daily life, it accounts for the largest proportion at 47\%. Optics and thermodynamics, which typically require specific temperature or environmental conditions, account for 29\% and 24\%, respectively. The proportions of each subcategory are shown in the outer ring of the figure. Based on the labels of the manually collected videos, we evaluate the accuracy of the qualitative physical category annotations. The results are shown in the Figure. \ref{fig:data_correct}, where the accuracy for dynamics, optics, and thermodynamics reaches 84\%, 71\%, and 64\%, respectively, with an overall accuracy of 75\%.

\section{More Examples and Annotation}
\label{example_annotation}
\begin{figure*}
  \centering
    \includegraphics[width=1.0\linewidth]{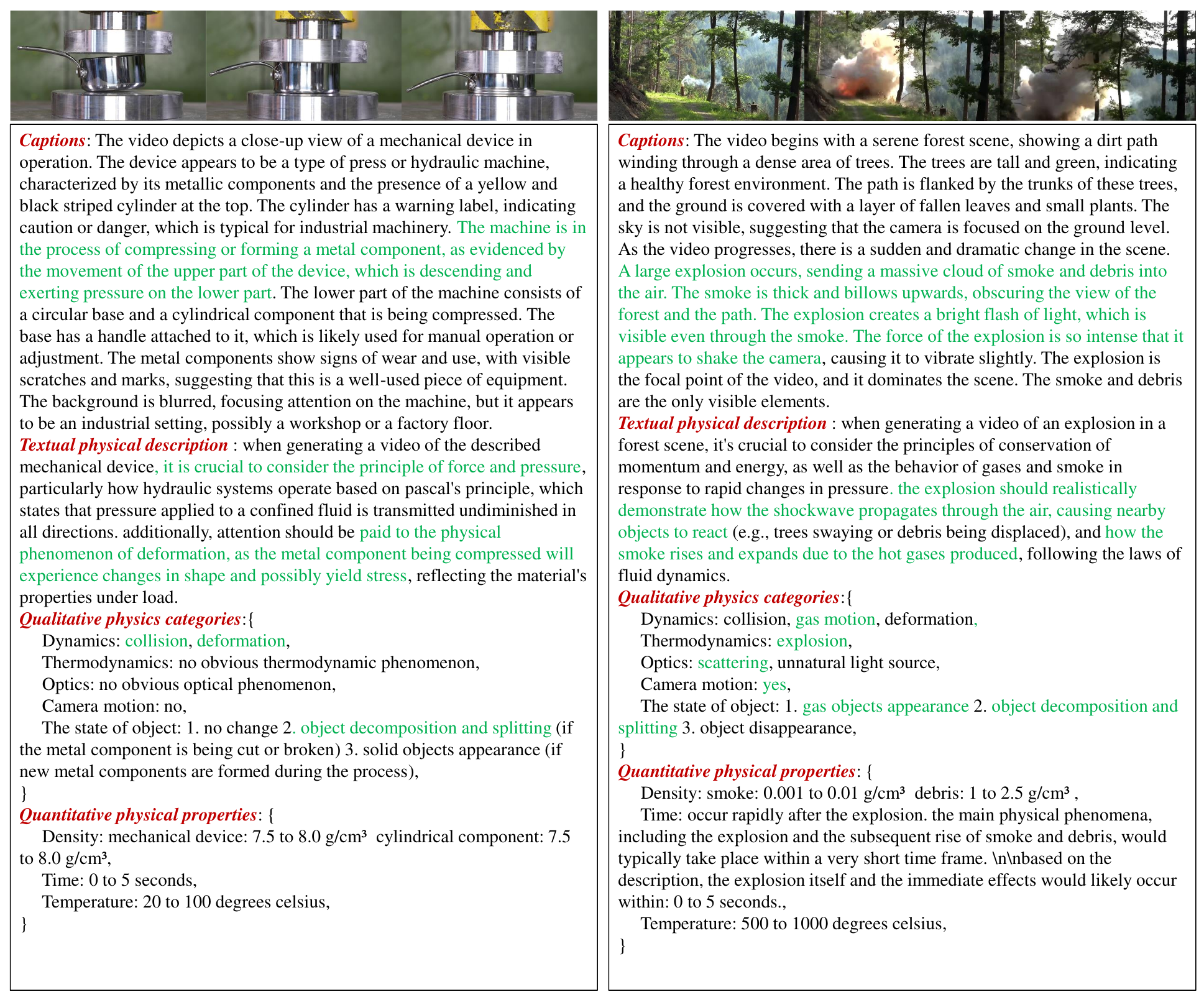}
  \caption{The video data and its detailed annotations in WISA-32K.}
  \label{fig:data_exsample}
\end{figure*}

Following the proposed physical information annotation pipeline, we construct the WISA-32K dataset. Several example videos and their corresponding annotations are shown in the Figure. \ref{fig:data_exsample}. This pipeline enables accurate and detailed annotation of physical information, ensuring that each video is comprehensively labeled with its relevant physical properties and phenomena.

\section{Annotation Prompts}
\label{annotation_prompts}
\begin{figure*}
  \centering
    \includegraphics[width=1.0\linewidth]{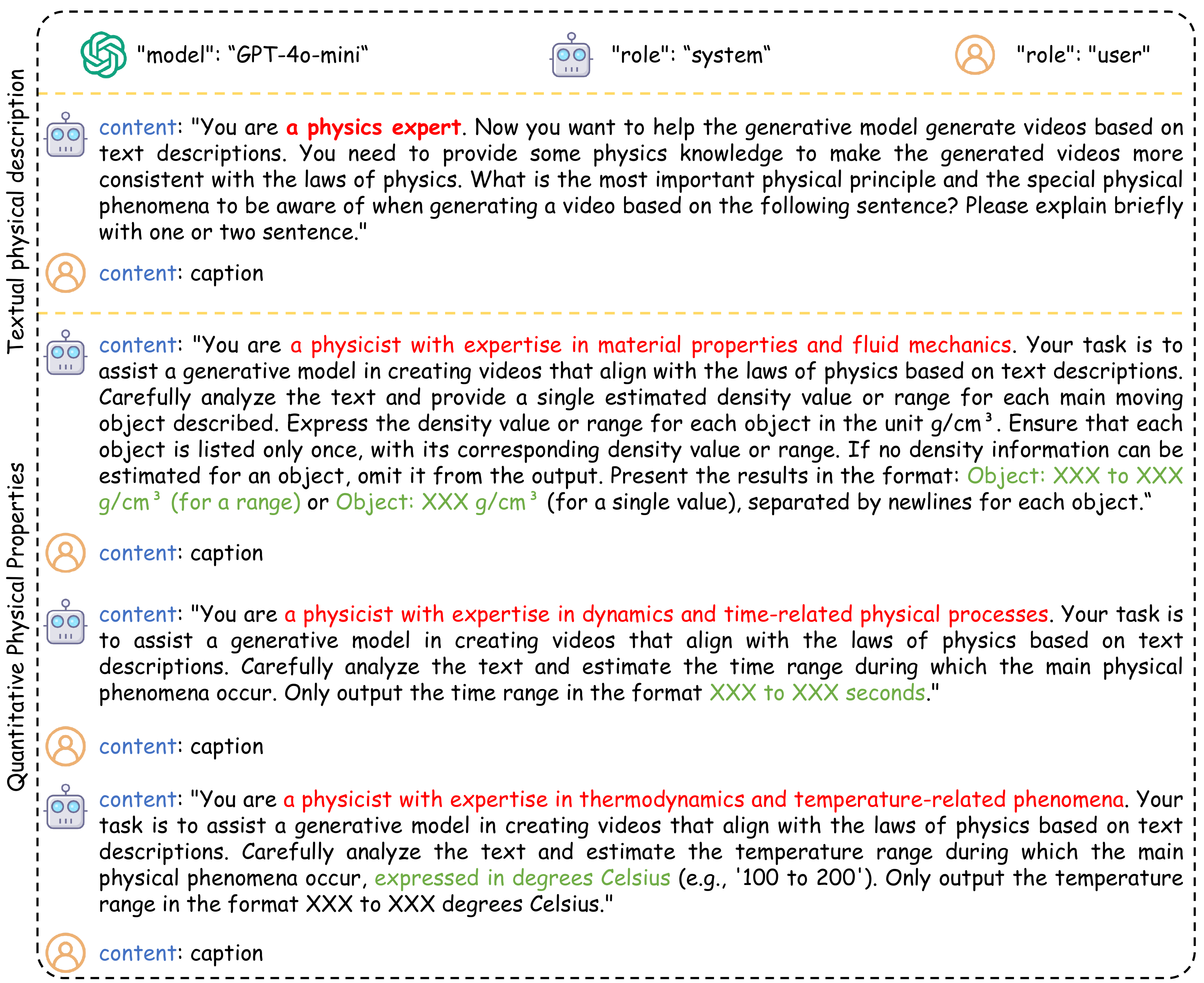}
  \caption{Prompts for annotating textual physical descriptions and quantitative physical properties}
  \label{fig:appendix_caption1}
\end{figure*}

\begin{figure*}
  \centering
    \includegraphics[width=1.0\linewidth]{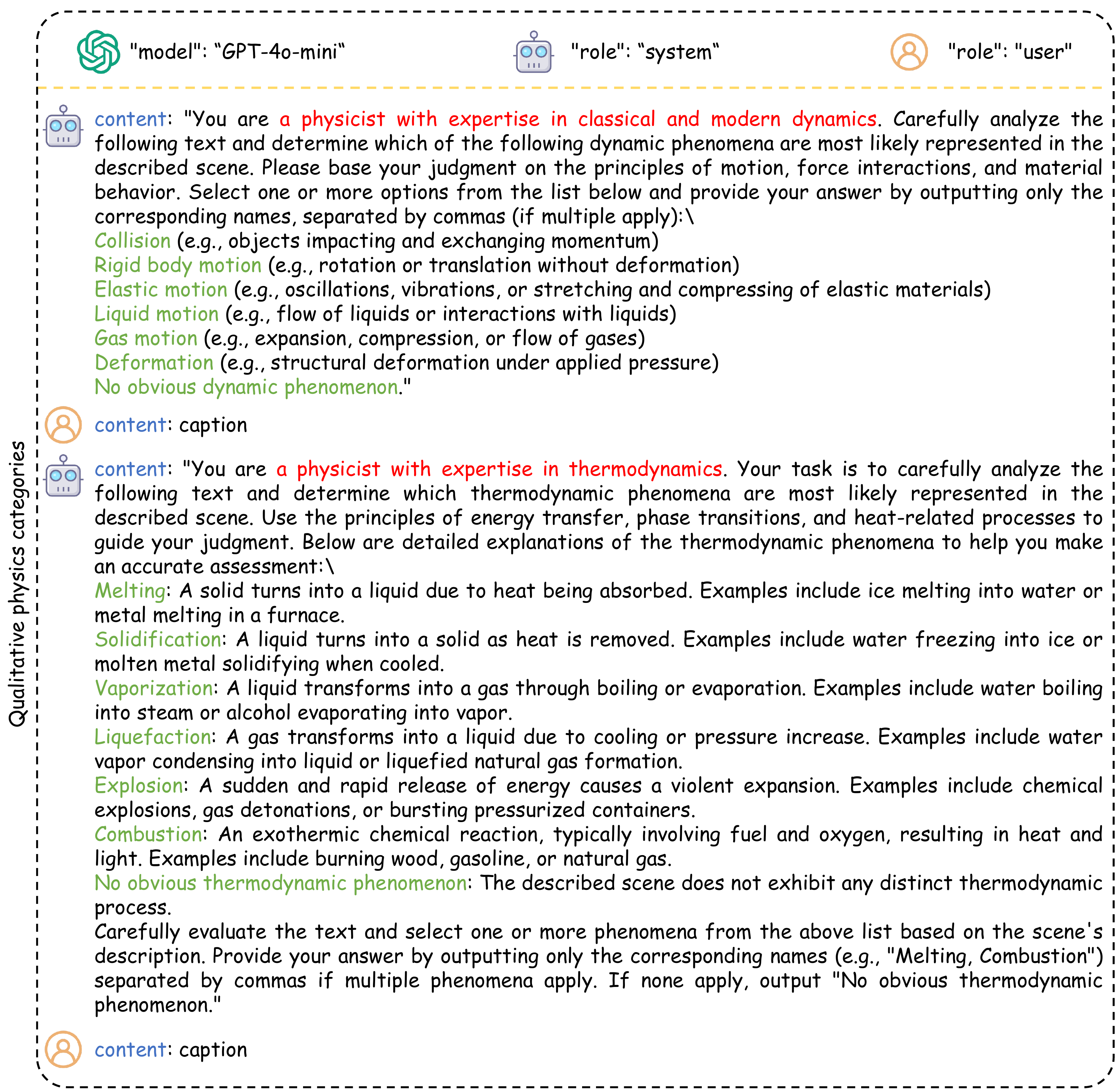}
  \caption{Prompts for annotating qualitative physics categories}
  \label{fig:appendix_caption2}
\end{figure*}

\begin{figure*}
  \centering
    \includegraphics[width=1.0\linewidth]{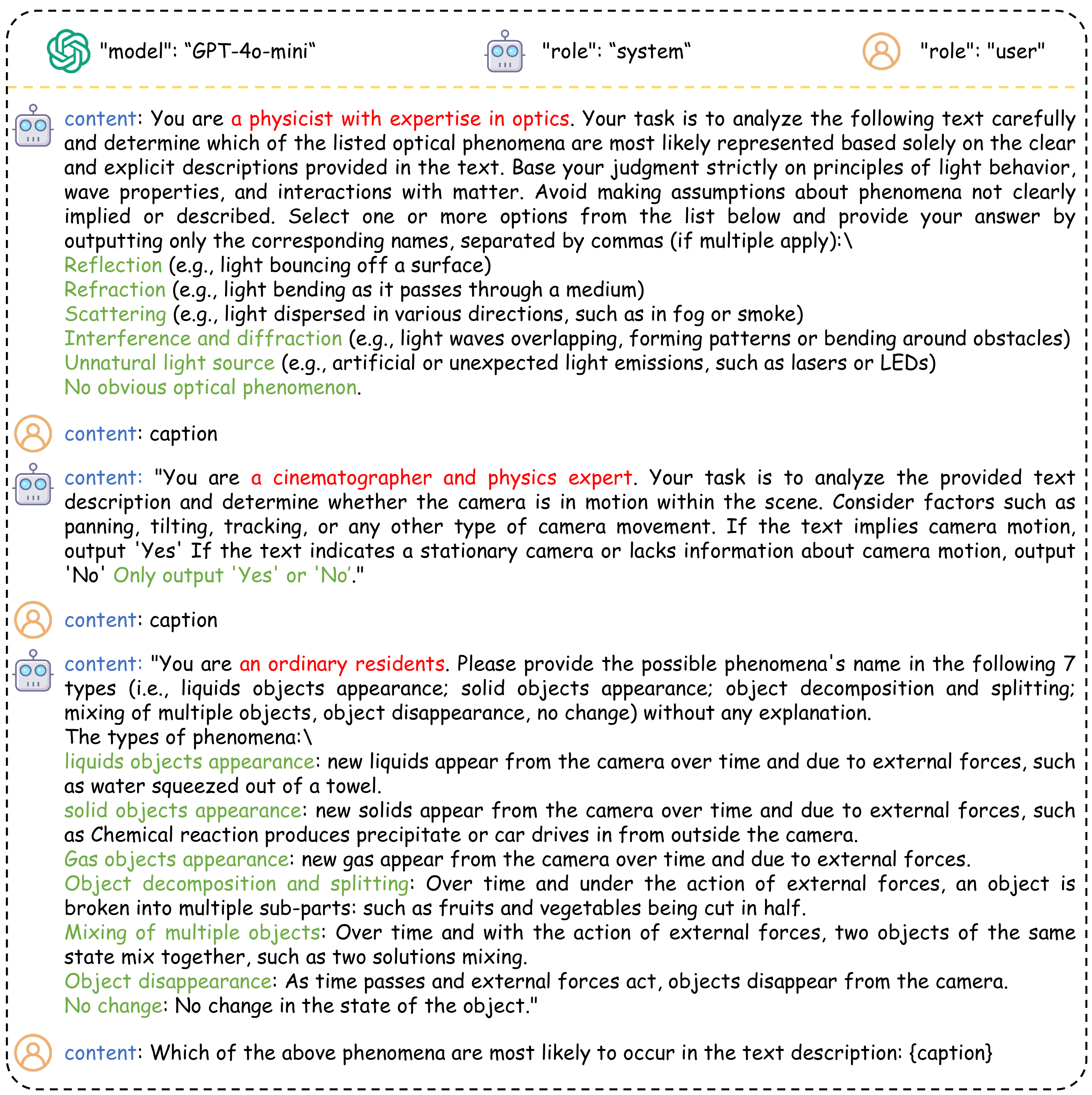}
  \caption{Prompts for annotating qualitative physics categories}
  \label{fig:appendix_caption3}
\end{figure*}

The detailed prompt used for physical information annotation is illustrated in the Figure. \ref{fig:appendix_caption1}, Figure. \ref{fig:appendix_caption2}, and Figure. \ref{fig:appendix_caption3}.

\section{Word Cloud}
\label{word_cloud}
\begin{figure*}
  \centering
    \includegraphics[width=1.0\linewidth]{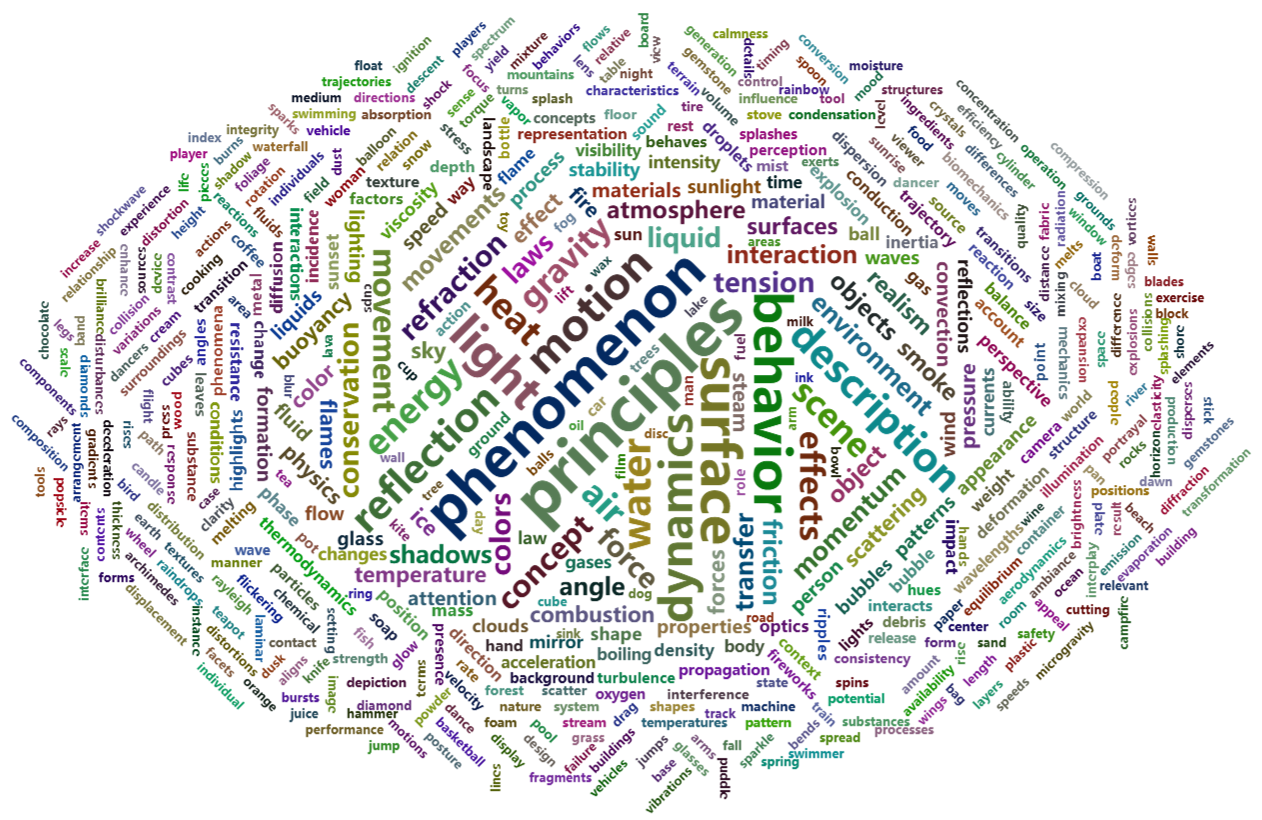}
  \caption{Word cloud generated from textual physical description, where larger words indicate higher frequencies in the dataset text}
  \label{fig:word_cloud}
\end{figure*}

We conducted a word frequency analysis on the textual physical description in the dataset and generated the word cloud shown in Figure. \ref{fig:word_cloud}. To filter out irrelevant words, we retained only nouns and selected them based on their frequency, from highest to lowest. Notably, physical terms such as 'motion,' 'phenomenon,' and 'light' appear more frequently, highlighting the strong physical relevance of the dataset.

\section{Discussion of Quantitative Evaluation}
\begin{figure*}
  \centering
    \includegraphics[width=1.0\linewidth]{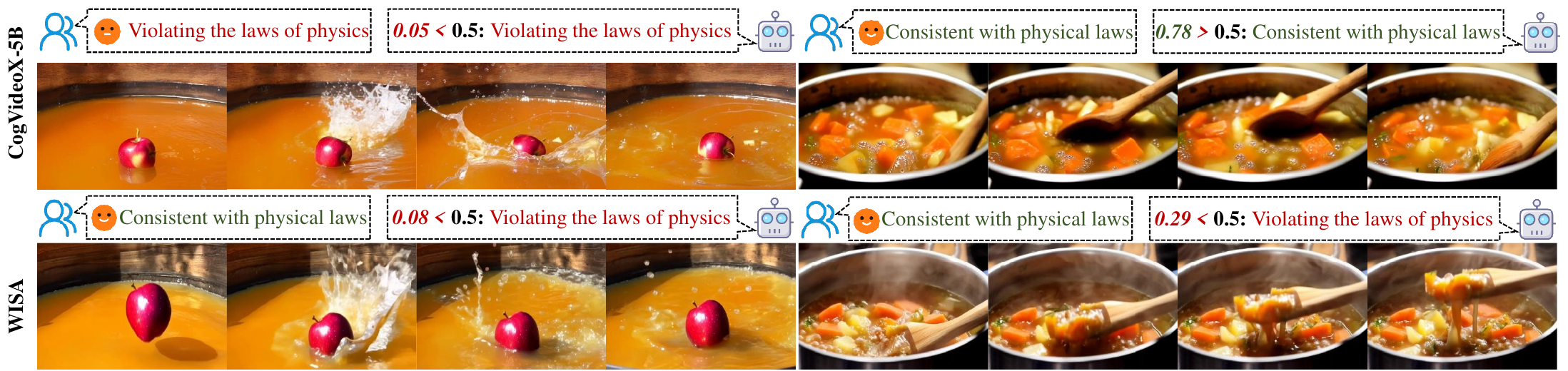}
  \caption{Human and machine evaluation results do not fully align.}
  \label{fig:sup_bench_error}
\end{figure*}

During the quantitative evaluation, we observe several misjudgments in VideoCon-Physics, as shown in the Figure. \ref{fig:sup_bench_error}. Specifically, WISA generates a physically plausible process where the object enters the water first, followed by the splash — aligning well with real-world physical laws. However, this sample only receives a low score of 0.08 from VideoCon-Physics.
We further conduct a simple test using Qwen2.5-VL for evaluation, and the model also struggles to distinguish the correct or incorrect sequence of physical events. These findings show the limitations of existing video-based physics evaluation metrics, indicating that future research into more reliable physical property assessments for videos is necessary.

\end{document}